\documentclass[runningheads]{llncs}

\usepackage{multirow}
\usepackage{graphicx}
\usepackage{tikz}
\usepackage{xcolor}
\usepackage{array}
\usepackage{makecell}
\usepackage{enumitem}
\usepackage[figuresright]{rotating}

\usepackage{tabularx}
\usepackage[hidelinks]{hyperref}
\usetikzlibrary{arrows}
\usepackage{float}

\usepackage{amsmath}
\usepackage{amssymb}
\usepackage{stackengine}
\usepackage{pgfplots}
\pgfplotsset{compat=newest}
\usepackage{subcaption}
\usepackage[linesnumbered,ruled,vlined]{algorithm2e}
\usepackage{pifont}

\newcommand{\TODO}[1]{~\\\noindent\textcolor{magenta}{\fbox{\textbf{TODO:} #1}}}	% general todo macro

\begin{document}

\title{Ontology-Based Process Modelling - \\Will we live to see it?}

%Repairing Declarative Process Models via Principles of Minimal Change
\author{Carl Corea\inst{1} \and Michael Fellmann\inst{2} \and Patrick Delfmann\inst{1}}
\authorrunning{C. Corea et al.}

\institute{University of Koblenz-Landau, Germany\\
\email{\{ccorea,delfmann\}@uni-koblenz.de}\\
\and
University of Rostock, Germany\\
\email{michael.fellmann@uni-rostock.de}}
\maketitle              
\begin{abstract}
In theory, ontology-based process modelling (OBPM) bares great potential to extend business process management. Many works have studied OBPM and are clear on the potential amenities, such as eliminating ambiguities or enabling advanced reasoning over company processes. However, despite this approval in academia, a widespread industry adoption is still nowhere to be seen. This can be mainly attributed to the fact, that it still requires high amounts of manual labour to initially create ontologies and annotations to process models. As long as these problems are not addressed, implementing OBPM seems unfeasible in practice. In this work, we therefore identify requirements needed for a successful implementation of OBPM and assess the current state of research w.r.t. these requirements. Our results indicate that the research progress for means to facilitate OBPM are still alarmingly low and there needs to be urgent work on extending existing approaches.

\end{abstract}

%==============================
%
%		    	Intro
%
%==============================
\section{Introduction}
In the scope of Business Process Management, \emph{process models} have evolved as central artifacts for the design, enactment and analysis of company processes \cite{weske2007business}. Many modelling languages, such as the Business Process Model and Notation\footnote{\url{https://www.omg.org/bpmn/}}, are available and have received widespread adoption in practice. While these standards offer support for the \emph{representation} of company processes, the actual \emph{content} of the model is still the responsibility of the modeller. That is, process models are designed by human modellers, often times also in a collaborative and incremental manner. In this setting, modelling errors can occur frequently \cite{riehle2017automatically,Fellmann:2015}. For example, humans might accidentally model a non-compliant sequence of activities or use ambiguous activity labels. %Also, as the activities are captured with natural-language descriptions (labels), different views or understandings of modellers can lead to terminological issues in the resulting models, e.g., ambiguities of the prescribed activities or duplicate elements that clutter the model.

\begin{figure}[]
    \centering
    \includegraphics[width=.5\textwidth]{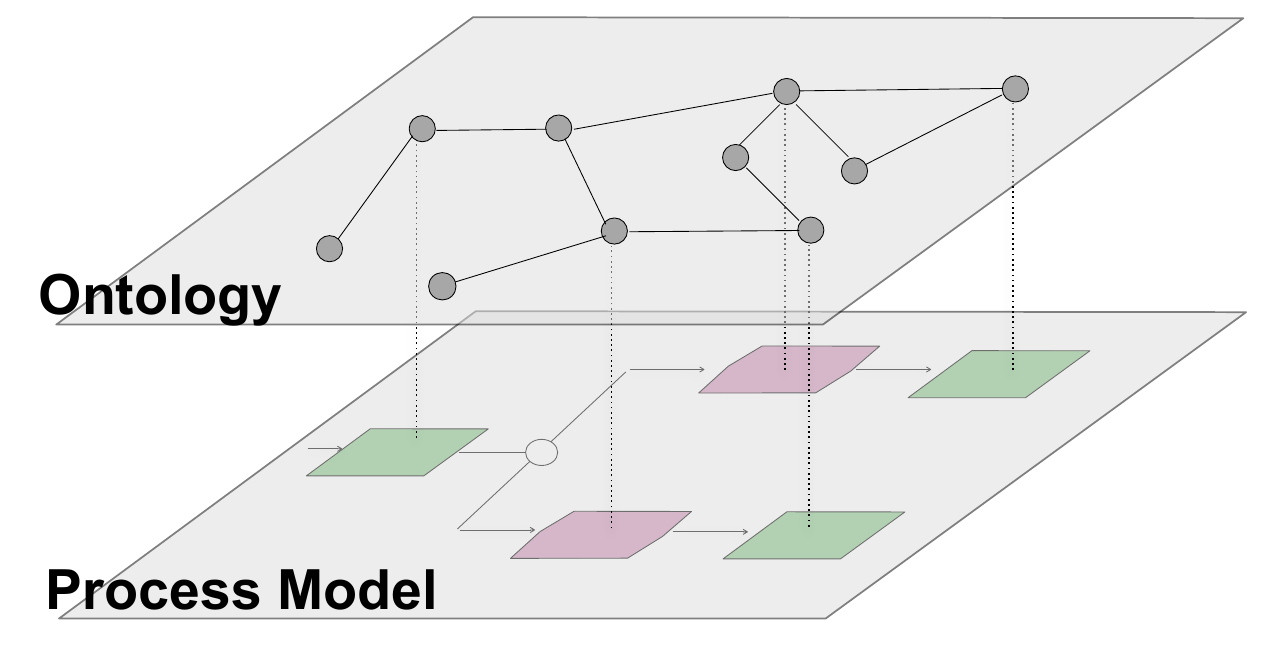}
    \caption{Exemplary Ontology-Based Process Model.}
    \label{fig:exemplary_model}
\end{figure}

To conquer such modelling problems, there is a broad consensus in academia that business process models should be extended with an additional conceptual layer, namely \emph{ontologies} \cite{Fellmann:2015}. Ontologies are artifacts that can be used to formally conceptualize a domain of interest \cite{corea2017detecting}. As shown in Figure \ref{fig:exemplary_model},  \emph{ontology-based process models} can thus be created by extending process models with ontologies, which allows to define unambiguous semantics of process models and create a shared understanding of business processes for humans and machines alike. To clarify, the ontologies addressed in this works are limited to models that are coded in description logics, e.g., OWL-ontologies linked to process models.

Many works have discussed the potential benefits of OBPM, e.g. eliminating ambiguous process descriptions or advanced compliance reasoning using ontology-based reasoning \cite{Fellmann:2015,corea2017detecting}. Still, industry adoptions are sparse. Based on reports such as \cite{riehle2017automatically}, this can be mainly attributed to the problems in creating the ontology-based process models themselves. First, ontologies must initially be created. As this requires a high expertise in knowledge representation, this is currently still a difficult task for companies. Second, even given an ontology, the ontology has to be annotated to the process model. As finding the connections between ontology- and process model elements is a complex task, manual annotation can be seen as highly unfeasible in practice \cite{riehle2017automatically}. %Third, annotated process models have to be kept in synchronization with the ontology, i.e., a co-evolution of ontologies and process models has to be supported. For example, if an ontology concept changes, some annotations may become invalid and have to be updated. 

While there have been works proposing initial means for the problems raised above, the lack of industry adoptions suggests that companies need more support in implementing OBPM. In this report, we therefore investigate what methods and results are still missing to support companies in implementing OBPM and leverage industry adoption. Here, our contributions are as follows: 
\begin{itemize}
    \item[\textbullet] \textbf{Requirements for OBPM.} We identify requirements needed for successfully implementing and maintaining ontology-based process models for companies. 
    \item[\textbullet] \textbf{Literature Analysis.} We identify the state-of-the-art of OBPM research based on a literature review and assess to which extent current results support the identified requirements.
    \item[\textbullet] \textbf{Research Agenda.} We identify current research gaps and distill a research agenda to guide future research and tool development.
\end{itemize}
This work is structured as follows. In Section 2, we provide preliminaries and raise requirements for implementing OBPM. In Section 3, we analyze the state-of-the-art on OBPM research and identify research gaps, from which we distill a corresponding research agenda in Section 4. We conclude in Section 5.

%==============================
%
%		   Preliminaries
%
%==============================
\section{Preliminaries}

%OBDA
\subsection{Ontology-Based Process Models}
Ontology-based process models \cite{thomas2009semantic} allow to define the semantics of business process model elements by extending traditional process models with an additional ontology-layer. To this aim, elements of business process models can be annotated to ontology concepts via mappings, also referred to as \emph{semantic annotation} \cite{fellmann2011checking,riehle2017automatically}. This creates a shared conceptual and terminological understanding and defines a machine-accessible semantics of company processes.

Figure 1 shows an exemplary ontology-based process model. A company ontology can be used to define important company knowledge, e.g., concepts such as organizational units, tasks, services or business rules. These ontology concepts can then be annotated to elements of the process model \cite{thomas2009semantic}. Technically, this is performed by creating axiomatical knowledge in the ontology, i.e., instances of ontology concepts, which each represent a process model element. In this way, the semantics and relations of the process model elements can be formalized.

Regarding ontology-based process models, the potential advantages are clear from an academic standpoint \cite{Fellmann:2015,riehle2017automatically,di2009supporting,leopold2015towards,hassen2017using,thomas2009semantic}. Such artifacts foster a shared semantic understanding for humans and machines and allow for advanced reasoning capabilities over process models for companies. Still, a widespread industry adoption cannot be observed. As this could be attributed to companies having problems implementing OBPM, we therefore investigate requirements of implementing OBPM in the following to gain an understanding about whether there could be any obstacles impeding industry adoption.

%Annotation
\subsection{Requirements for implementing OBPM}\label{sec:requirements}
On a high-level, implementing OBPM can be distinguished into ensuring three components: \emph{an ontology, a process model} and \emph{a mapping between the ontology and the process model}. This might seem trivial, however, to anticipate the results of our literature analysis, there was no approach that supported companies in creating and maintaining all three of these components. For example, virtually all existing approaches assume ``ready-to-go" ontologies that can be used as a basis for extending process models. Here, following works such as \cite{gailly2017recommendation}, we argue that this assumption might not be plausible in an enterprise setting, an rather, companies must also be supported in the initial creation of ontologies (R1). Also, following woks such as \cite{riehle2017automatically}, both the ontology- and process model labels should be initially standardized, as a basis for extending process models (R2). 

Furthermore, companies must be able to create mapping of ontologies and process models. Here, there is a broad consensus that companies should be supported by means of mapping- and annotation techniques (R3) \cite{riehle2017automatically,di2009supporting}. 

\begin{figure}[H]
    \centering
    \includegraphics[width=\textwidth]{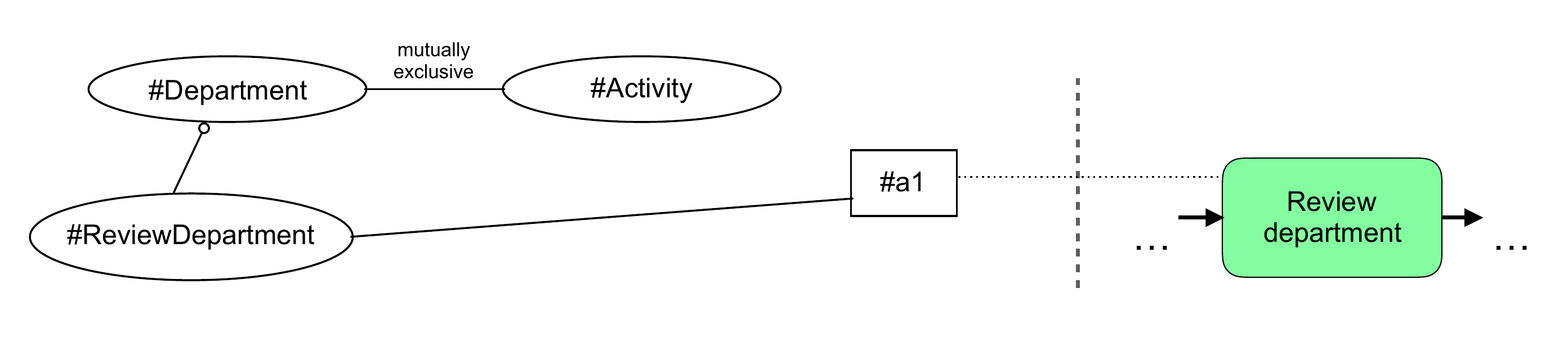}
    \caption{Exemplary linguistic-based annotation leading to incoherent mappings.}
    \label{fig:inconsistencyExample}
\end{figure}

When mappings are created, these mappings between ontology concepts and process model element have to be checked for plausibility. %That is, as opposed to "simply" checking whether certain elements of the ontology, such as organizational units, are \emph{missing} in the process model (which is a common use-case of queries against against annotated process models), it has to be verified whether the artifacts contain logically \emph{inconsistent} specifications, i.e., contradictory specifications that cannot hold at the same time. 
A problem here is that the mappings are mostly created based on terminological similarity between ontology concepts and process model elements. This however means that the semantic contexts of these elements are mostly disregarded, which in turn can lead to implausible mappings. For instance, consider the example in Figure \ref{fig:inconsistencyExample}. On the ontology level, the company has defined departments and activities, both of which are mutually exclusive. Furthermore, we see a specialized review department. Given a linguistic-based mapping approach matches the ontology concept \emph{Review department} with the activity \emph{Review department} (this would be highly sensible based on various linguistic-based similarity measures), this would result in a case where a process element is both an activity and a department, which, as defined in the ontology, is not possible. 
\vspace{-.5cm}
\vspace{-.7cm}
Thus, while the mapping might be sensible from a terminological viewpoint, from a global perspective, the ontology-based process model is inconsistent, i.e., it cannot be instantiated. In the field of ontology matching, this problem is referred to as \emph{mapping incoherence} (i.e., mappings that induce inconsistencies) and is widely acknowledged as a challenging task \cite{meilicke2008incoherence}. %, which can also occur when mapping ontologies and process models without sufficiently integrating the respective semantic contexts. %Therefore, such problems must be resolved in order to even be able to use the models. %Next to mappings that induce inconsistencies, 
Further challenges are also mappings that are simply implausible in a real-life sense, e.g., mapping a process model activity from a sales process to an ontology concept describing a purchase process if they have similar labels. An annotation verification is therefore necessary during modelling to ensure the plausibility of annotations (R4).

Soundness checks are especially relevant in the scope of model evolution. For instance, given an annotated process model, if either the underlying ontology or process is edited, the corresponding counterpart should be altered as well to ensure consistent co-evolution. Here, companies need to be supported in co-evolving ontologies and process models in a synchronized manner through an automatic application of change patterns, e.g. deleting something in the ontology if a corresponding element is deleted in the process model (R5).

Only if the requirements R1-R5 are met, the ontology-based process model can and should be used for ontology-based reasoning. For example, the process model can be verified for compliance by means of ontology-based reasoning (R6).

Withal, an important requirement of implementing OBPM is that the capabilities w.r.t. R1-R6 have to be tailored to the skill set of the stakeholders involved in the respective companies' business process managament. Otherwise, it cannot be ensured that the technical capabilities can even be used correctly. Therefore, an important aspect of implementing OBPM are means for assessing the actual usability of the underlying technical capabilities (R7).

Consequently, we raise the following requirements for implementing OBPM:

\begin{itemize}

    \item[\textbf{R1}.] \textbf{Ontology Support:} Support to create ontologies.
    \item[\textbf{R2}.] \textbf{Terminological Standardization:} Support to standardize ontologies and process models.
    \item[\textbf{R3}.] \textbf{Annotation Support:} Support to create semantic annotations.
    \item[\textbf{R4}.] \textbf{Annotation Verification:} Support to verify soundness (consistency, plausibility) of annotations. 
    \item[\textbf{R5}.] \textbf{Evolution Support:} Support to further evolve and maintain ontology-based models (in a synchronized manner).
    \item[\textbf{R6}.] \textbf{Ontology-based reasoning:} Support to exploit the ontology-based process model, e.g., reasoning about process model, ontology-based recommendations during modelling.
    \item[\textbf{R7}.] \textbf{Use Case and Contextual Fit:} The implemented means for OBPM must be aligned with the skill set of the involved stakeholders. Also, OBPM results must actually offer competitive advantages, e.g., increased efficiency, increased understanding accuracy, or less mental effort needed for humans.
\end{itemize}

Our requirements catalogue is an extension of \cite{wetzstein2007semantic} who raise initial requirements regarding the implementation of modelling and analysis in OBPM (here: R5 and R6). As discussed, all the identified requirements must be met, as otherwise, it can be expected that OBPM can not be successfully implemented (e.g., the OBPM cannot be created without an ontology, etc.). %(e.g., the OBPM cannot be created without an ontology; the OBPM cannot be annotated if the ontology and the model labels are not standardized; the OBPM cannot be used for reasoning if the OBPM/annotations are inconsistent; OBPM reasoning cannot provide the expected value if the skillset of the involved stakeholders is not sufficient to work with the provided means; etc.). 
In the following, we therefore assess the state-of-the-art on OBPM research, in order to verify whether there exist sufficient means to meet these requirements.

%==============================
%
%		   Analysis
%
%==============================
\section{OBPM ``State-Of-The-Art" Analysis}\label{sec:analysis}
In this section, we present our research method and literature analysis.

%Method
\subsection{Method}
To assess the current state-of-the-art on OBPM research w.r.t. the requirements proposed in Section \ref{sec:requirements}, we conducted a literature review following the standard literature review method as proposed in \cite{vom2015standing}. First, we conducted brief searches to gain an initial overview, following the suggestion in \cite{rowley2004conducting}. %Here, we could identify initial seminal works, e.g., \cite{Fellmann:2015,fellmann2011checking}. 
We then used concept-mapping techniques to derive suitable keywords for a keyword-based literature search as proposed by \cite{vom2015standing}. In result, we utilized the keyword combination of ``Ontolog*" AND ``Process Model*", and ``Semantic" AND ``Process Model*". In iterative pre-testing, we found that these keywords are suitable to retrieve a substantial amount of relevant work. We queried 6 pertinent literature databases, i.e., ACM digital library, AISel, IEEE, Emeralds Publishing, INFORMS Pubs, Science Direct and Springer Link. This resulted in a total of 434 identified works. We then conducted a screening process in two review phases, as suggested by \cite{moher2014}. In review phase I, we eliminated duplicates and filtered out irrelevant articles based on their title and abstract. Then, in review phase II, all remaining articles were read in full and assessed for relevance. Here, we defined an article to be relevant if its primary focus is OBPM or an intersection of BPM and OBPM. Also, due to funding constraints, the article had to be publically available to be included. Based on these review phases, we identified 47 works as relevant. For the works selected to be relevant, we further performed a backward- and forward-search (by references), as suggested in \cite{webster2002analyzing}. %which resulted in 3 identified relevant works. %For backward search, we conducted a backward search by references, where we regarded all the sources referenced by the initially found relevant results. For a forward search, we conducted a forward search via Web of Science to identify works that cited the relevant results. The works identified via the forward- and backward-search were assessed by the same criteria as applied in review phase I and II. 
This resulted in 4 additional identified works. To summarize, we could identify 51 (47 keyword-based + 4 f/b) relevant works by means of our literature review. In the following, we assess these works in regard to our proposed requirements.

%SOTA
\subsection{Literature Analysis}
Table \ref{tab:literature_analysis} shows which of the raised requirements R1-R7 is satisfied (x), resp. partially satisfied (o), by the identified relevant works. 

\begin{table}[H]
    	    
	\scriptsize
	\hskip-1.25cm
	\begin{tabular}{|l||c|c|c|c|c|c|c|}
		\hline
		Work  & R1 & R2& R3 & R4& R5& R6 & R7 \\
		\hline
		Lin  et al. (2005)\cite{lin2005ontology}                  &   &   & o   &   &   & o &\\
		\hline
		Thomas et al. (2005)\cite{thomas2005semantic}             &   &   & o   &   &   & o &\\
		\hline
		Lautenbacher et al. (2006)\cite{lautenbacher2006semantic} &   &   & o   &   &   & o &\\
		\hline
		Ma et al. (2006)\cite{ma2006research}                     &   &   & o   &   &   & o &\\
		\hline
		Barjis et al. (2007)\cite{barjis2007executable}           &   &   &     &   &   & x &\\
		\hline
		Born et al. (2007)\cite{born2007user}                     &   & x &  x   &   &   &   &\\
		\hline
		Ehrig et al. (2007)\cite{Ehrig2007}                       &   &   & o   &   &   &   &\\
		\hline
		Höfferer et al. (2007)\cite{hofferer2007achieving}        &   &   & x   &   &   &   &\\
		\hline
		Sujanto et al. (2008)\cite{sujanto2008application}        &   &   &     &   &   & x &\\
		\hline
		Cabral et al. (2009)\cite{cabral2009business}             &   &   & o   &   &   & o &   \\
		\hline
		Di Franc. et al. (2009)\cite{di2009supporting}   &   &   & x   &   &   &   &   \\
		\hline
		Fengel et al. (2009)\cite{fengel2009model}                & x &   & x   &   &   &   &   \\
		\hline
		Hinge et al. (2009)\cite{hinge2009process}                &   &   & o   &   &   &   &  \\
		\hline
		Markovic et al. (2009)\cite{markovic2009process}          &   &   &     &   &   & x &   \\
		\hline
		Norton et al. (2009)\cite{norton2009ontology}             &   &   & o   &   &   &   & \\
		\hline
		Thomas et al. (2009)\cite{thomas2009semantic}             &   &   & o/m &   &   & x &   \\       
		\hline
		Barnickel et al. (2010)\cite{barnickel2010incorporating}  &   &   & m   &   &   & x &   \\
		\hline
		Hua et al. (2010)\cite{hua2010exploring}                  &   &   &     &   & o & o &   \\
		\hline
		Tan et al. (2010)\cite{tan2010approach}                   &   &   & o   &   & o & o &  \\
		\hline
		Benaben et al. (2011)\cite{benaben2011collaborative}      &   &   &     &   &   & x &   \\
		\hline
		Fellmann et al. (2011)\cite{fellmann2011checking}         &   &   & o/m &   &   & x &   \\
		\hline 
		Fellmann et al. (2011)\cite{fellmann2011query}         &   &   & &   &   & x & x  \\
		\hline 
		Kim et al. (2011)\cite{kim2011semantic}                   &   &   & o   &   &   & o &   \\
		\hline
		Gong et al. (2012)\cite{gong2012executability}            &   &   &     & x &   &   &   \\
		\hline
		Stolfa et al. (2012)\cite{vstolfa2012business}            &   &   & o   &   &   &   &   \\
		\hline
		Fill et al. (2013)\cite{fill2013generic}                  &   &   & o   &   &   &   &   \\
		\hline     
	\end{tabular}
	\quad
	\begin{tabular}{|l||c|c|c|c|c|c|c|}
		\hline
		Work  & R1 & R2& R3 & R4& R5& R6 & R7 \\
		\hline       
		Kherbouche et al. (2013)\cite{kherbouche2013ontology}       &   &   &     &   & o & o &                                           \\
		\hline
		De Cesare et al. (2014)\cite{de2014toward}                  & x &   &     &   &   &   &                                                      \\
		\hline
		Maalouf et al. (2014)\cite{maalouf2014semantic}             &   &   & o   &   & o & o &                                                 \\
		\hline
		Rospocher et al. (2014)\cite{rospocher2014ontology}         &   &   & o   &   &   & o &                                                     \\
		\hline
		Wang et al. (2014)\cite{wang2014context}                    &   &   & o   &   &   & o &                                                      \\
		\hline
		Di Martino  et al. (2015)\cite{di2015semantic}              &   &   &     &   &   & x &                                                                      \\
		\hline
		Kalogeraki et al. (2015)\cite{kalogeraki2015semantic}       &   &   &     &   &   & x &                                                            \\
		\hline
		Leopold et al. (2015)\cite{leopold2015towards}              &   &   & x   & o  &   &   &                                   \\
		\hline
		Ngo et al. (2015)\cite{ngo2015using}                        &   &   & o   &   & o & o &                                              \\
		\hline
		Fan et al. (2016)\cite{fan2016process}                      &   & x & o   & x &   & o & x  \\
		\hline
		Pham et al. (2016)\cite{pham2016ontology}                   &   &   &     &   &   & x &                                                                           \\
		\hline
		Corea et al. (2017)\cite{corea2017detecting}                &   &   &     &   &   & x &                                                            \\
		\hline       
		Effendi et al. (2017)\cite{effendi2017swrl}                 &   &   &     &   &   & x &                                                          \\
		\hline
		Elstermann et al. (2017)\cite{elstermann2017proposal}       &   &   & o   &   &   & o &                                                       \\
		\hline       
		Fauzan et al. (2017)\cite{fauzan2017structure}              &   &   & x   &   &   & x &                                                                         \\
		\hline
		Gailly et al. (2017)\cite{gailly2017recommendation}         &   &   &     &   & x &   &                                  \\
		\hline           
		Hassen et al. (2017)\cite{hassen2017using}                  &   &   & m   &   &   & x &                     \\
		\hline
		Pawelozek et al. (2017)\cite{paweloszek2017process}         &   &   & m   &   &   & x &                                           \\
		\hline
		Riehle et al. (2017)\cite{riehle2017automatically}          &   & x & x   &   &   &   &                                         \\
		\hline
		Soltysik et al. (2017)\cite{soltysik2017organizational}     &   &   & m   &   &   & x &                                                       \\
		\hline
		Sungoko et al. (2017)\cite{sungkono2017refining}            &   &   &     & x & x &   &                               \\
		\hline	
		Bartolini et al. (2019)\cite{bartolini2019enhancing}        &   &   & o   &   &   & x &                                         \\
		\hline
		Wang et al. (2019)\cite{wang2019ontology}                   &   &   & o/m &   &   & o &                                                                       \\
		\hline
		Yanuarifiani et al. (2019)\cite{yanuarifiani2019automating} &   &   & o   &   &   & o &                                                                          \\
		\hline
		Cao et al. (2021)\cite{cao2021ontology}                     & x &   &     &   &   &   &                                                   \\
		\hline
	\end{tabular}
   
	\caption{Support of the considered works w.r.t. the identified requirements for implementing OBPM (x = satisfied; o = partially satisfied; m = manual).}
	\label{tab:literature_analysis}
	
\end{table}

\begin{figure}
    \centering
    \includegraphics[width=.8\textwidth]{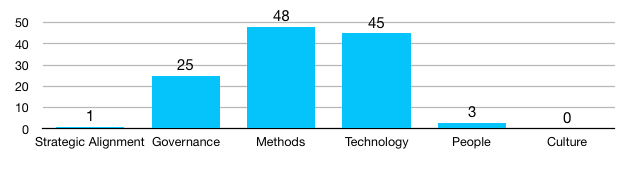}
    \caption{Focus of the identified works (cf. Table \ref{tab:literature_analysis}) in the scope of the six-core-elements of BPM framework \cite{rosemann2015six} (Multiple counting possible).}
    \label{fig:six_core}
\end{figure}

From the identified literature, it can be observed that the level of research in regard to 5 out of the 7 raised requirements is alarmingly low (R1, R2, R4, R5, R7). Also, regarding R3, many of the works only partially satisfy this requirement due to different limitations (cf. the below discussion). In result, we see the following research gaps:

\begin{itemize}
    \item[\textbullet] \textbf{Insufficient ontology support.} Virtually all identified approaches assume ``ready-to-go" ontologies for an extension of process models. This places a heavy burden on companies for adopting these approaches, as initially creating the ontology is left at the company's responsibility. Here, novel approaches are needed that integrate ontology creation in the OBPM-lifecycle, e.g., ontology-mining from textual descriptions, model repositories or process data (event logs).
    \item[\textbullet] \textbf{Limitations in annotation support.} Creating semantic annotations can be seen as one of the most important components of implementing OBPM. For creating semantic annotations, we can identify three main approaches from the identified literature, namely \emph{manual annotation, transformation,} and \emph{matching}. 1) Manual annotation (denoted $m$ in Table \ref{tab:literature_analysis}) is geared towards creating annotations manually. While we agree that this could be used for revision or fine-tuning, a completely manual annotation of process models must be considered as unfeasible, as this forces modelers to compare thousands of elements and map their relations manually \cite{riehle2017automatically}. 2) Approaches with annotation through transformation (denoted $o$ in Table \ref{tab:literature_analysis}) implement an automated annotation by transforming every process model element into an ontology instance. While this creates a representation of the process model on the ontology level, a major limitation of this approach is that the annotations are on a pure syntactical level, i.e., process model elements are instantiated as an instance of their type, however, the ``real-life" semantics of the process activities cannot be captured. 3) To facilitate such a form of ``semantic" annotations, a third line of approaches proposes means to match process model element and ontology concept semantics by matching semantically equivalent model and ontology elements, mainly by means of linguistic analysis (denoted x in Table \ref{tab:literature_analysis}). % In summary, manual approaches or approaches for creating ``simple" syntactic annotations cannot be seen as sufficient for the abovementioned reasons. 
    Regarding this line of approaches, while see some results based on terminological matching, we see a strong lack of works integrating and/or combining other matching techniques, e.g., constraint-based matching, graph-based matching, instance (data) based matching, model based matching or incorporating context information in matching (cf. e.g. the classification of matching techniques in \cite{otero2015ontology}). Here, more approaches and comparative studies are needed to facilitate better automated annotation. Also, supervised machine-learning approaches are needed that can mimic human annotations. 
    \item[\textbullet] \textbf{Missing verification support.} As motivated in Section 2, a natural language based matching between process model elements and ontology elements without considering their contexts can lead to the resulting ontology-based process model being inconsistent or implausible. Yet, the majority of annotation approaches via matching implement such an annotation based on linguistic mapping (cf. \cite{born2007user,fengel2009model,hofferer2007achieving,fauzan2017structure,di2009supporting,leopold2015towards,riehle2017automatically}). While this raises strong demand for annotation verification, we see from Table \ref{tab:literature_analysis} that there is currently no sufficient support for verifying consistency process model and the ontology. %Thus, even if elements can be matched via a linguistic matching, it can currently not be verified whether the matched artifacts are even consistent, i.e., whether the ontology-based process model can be instantiated. 
    Here, novel approaches are needed for analyzing and resolving inconsistencies in ontology-based process models.
    \item[\textbullet] \textbf{Missing evolution support.} Current works for annotations are mainly geared towards an \emph{initialization} of an ontology-based process model. However, as ontologies or the process models may need to be evolved over time, the other corresponding artifact also needs to be altered in a synchronized manner. Here, new methods are needed that can support companies in model evolution. Otherwise, there is no sufficient support for sustainable OBPM.
    \item[\textbullet] \textbf{Missing studies with human participants.} As a major warning-sign, we found almost no studies investigating the use of OBPM with human participants. Therefore, the effects for workers of introducing OBPM are unclear, e.g., do ontology-based process modelling really improve understanding and efficiency, or are they too complicated to use? Such experimental research should urgently be conducted. This also becomes apparent when assessing the considered works of our literature analysis in the scope of the \emph{core elements of BPM-framework} by \cite{rosemann2015six}. This framework identifies six core elements of business process management, namely \emph{strategic alignment, governance, methods, technologies, people} and \emph{culture}. As can be seen in Figure \ref{fig:six_core}, the identified works are mainly geared towards the dimensions of concrete methods and technologies. Here, we see a clear lack of research regarding OBPM relative to the skill sets of the involved people, organizational culture and structures.
\end{itemize}

%==============================
%
%		   Research Agenda
%
%==============================
\vspace{-.4cm}
\section{Research Agenda}
 %In this section, we have presented a procedure model for initially implementing an ontology-based process model. 
 In regard to the results of our literature analysis, it seems that much research is still needed to implement the requirements R1-R7 identified in this work. Based on the identified research gaps, we therefore propose the following research agenda to guide future research and leverage the implementation of OBPM, shown in Table \ref{tab:technical_research_agenda}

\begin{table}[H]
    \centering
    \scriptsize
    \begin{tabular}{p{3cm}|p{9cm}}
    
    \Xhline{2\arrayrulewidth}
    \textbf{Research Gap} & \textbf{Research needed}\\
    \Xhline{2\arrayrulewidth}
    %Ontology Support
    \vspace{1pt}
    \emph{Ontology Support} & 
        \begin{itemize}[noitemsep,nolistsep,leftmargin=*]
            \item[\textbullet] Integrating ontology creation in the OBPM-lifecycle
            \item[\textbullet] Ontology mining (e.g. from data, textual descriptions)
            \item[\textbullet] Recommendations during ontology-modelling
        \end{itemize}\\
        \hline
    %Standardization
    \vspace{1pt}
    %R2 - \newline\emph{(Terminological Standardization)} & 
        %\begin{itemize}[noitemsep,nolistsep,leftmargin=*]
            %\item[\textbullet] Enforcing terminological standardization during (ontology) modelling
            %\item[\textbullet] Standardizing ontologies (ex post)
            %\item[\textbullet] Managing and evolving glossars and domain thesauri during modelling
        %\end{itemize}\\
        %\hline    
    %Annotation
    \vspace{1pt}
    \emph{Annotation Support} & 
        \begin{itemize}[noitemsep,nolistsep,leftmargin=*]
            \item[\textbullet] Improving mapping approaches, e.g., with novel matching techniques (cf. the classification in \cite{otero2015ontology}) and novel similarity measures
            \item[\textbullet] Combining matching techniques to leverage matching
            \item[\textbullet] In case of repositories of ontology-based process models: Identify similarities of models in order to recommend annotations, e.g. via collaborative filtering, cf. \cite{chan2012assisting}
            \item[\textbullet] Creating AI systems that can learn/mimic human annotations
            \item[\textbullet] Consideration of context information for annotation suggestions (e.g. preceding and succeeding elements, granularity of surrounding model contents, general topic of the model, industry-specific process knowledge)
            \item[\textbullet] For evaluating matching techniques: Evaluation based on probabilistic gold standards \cite{leopold2012probabilistic} (i.e. averaged over multiple users), to avoid bias of the creator of the gold standard
        \end{itemize}\\
        \hline        
    %Verification Support
    \vspace{1pt}
    \emph{Annotation Verification} & 
        \begin{itemize}[noitemsep,nolistsep,leftmargin=*]
            \item[\textbullet] Identifying incoherent or implausible mappings between process models and ontologies
            \item[\textbullet] Analyzing inconsistencies or pin-pointing highly problematic elements, e.g. by means of element-based inconsistency measures
            \item[\textbullet] Means for resolving inconsistencies, e.g. in the scope of systems for semi-automated/guided inconsistency resolution
        \end{itemize}\\
        \hline   
    %Evolution
    \vspace{1pt}
    \emph{Evolution Support} & 
        \begin{itemize}[noitemsep,nolistsep,leftmargin=*]
            \item[\textbullet] Support of change propagation during modelling, e.g., automatedly synchronizing process model and ontology if an element is deleted in one of the two artifacts
            \item[\textbullet] Using ontology-based reasoning and insights from process monitoring to recommend changes needed in the process models or ontologies
            \item[\textbullet] Leveraging insights from process monitoring to recommend changes needed for the company ontology
            \item[\textbullet] Means for long-term model management, e.g. versioning
        \end{itemize}\\
        \hline  
        %Standardization
    \vspace{1pt}
    %R6 - \newline\emph{(Ontology-based reasoning)} & 
        %\begin{itemize}[noitemsep,nolistsep,leftmargin=*]
            %\item[\textbullet] Evaluation of approaches and tools for OBPM with human participants, i.e., do all the studied results really offer competitive advantages, for example in the form of increased efficiency, understanding accuracy or reduced cognitive load.
        %\end{itemize}\\
        %\hline    
    %User
    \vspace{1pt}
    \emph{Contextual Fit/People} & 
        \begin{itemize}[noitemsep,nolistsep,leftmargin=*]
            \item[\textbullet] Evaluation of approaches and tools for OBPM with human participants, i.e., do all the studied results really offer competitive advantages, for example in the form of increased efficiency, understanding accuracy or reduced cognitive load
            \item[\textbullet] Guidelines for the design of OBPM tools and interfaces
            \item[\textbullet] More research on how OBPM affects people and company culture (cf. Figure \ref{fig:six_core})
        \end{itemize}\\
        \hline   
    %Other
    \vspace{1pt}
    \emph{Other} & 
        \begin{itemize}[noitemsep,nolistsep,leftmargin=*]
            \item[\textbullet] Maturity models and procedure models for guiding companies in adopting OBPM
            \item[\textbullet] Success stories/case studies to share knowledge and leverage OBPM adoption
            \item[\textbullet] OBPM lifecycle models, e.g., how to integrate insights from process monitoring and ontology-based reasoning to improve process models and ontologies
            \item[\textbullet] Monitoring process execution for ontology-based process execution
        \end{itemize}\\
        \hline    
    \Xhline{2\arrayrulewidth}
    \end{tabular}
    \caption{Proposed Research Agenda}
    \label{tab:technical_research_agenda}
\end{table}

From a technical perspective, the distilled research agenda identifies numerous technical challenges that need to be addressed in regard to implementing OBPM, e.g., in the areas of \emph{ontology-support, annotation support, annotation verification} and \emph{evolution support}. Furthermore, research investigating the \emph{contextual fit} of OBPM is much needed. Here, on the one hand, there is work needed to assessing the added value of OBPM. On the other hand, the (cognitive) effects of introducing OBPM are still unclear, e.g., how can OBPM results be aligned with the skill sets of the involved stakeholders and does OBPM really offer positive cognitive effects for humans, such as better understanding efficiency? %Last, as the current state-of-the-art is very scattered (i.e., as shown in Table \ref{tab:literature_analysis}, the identified works address mostly only \emph{individual} requirements) work is needed on how to \emph{combine} existing results to make OBPM accessible for companies, e.g., lifecycle- or procedure models. %To this aim, we have proposed a first procedure model for the initial creation of an ontology-based process model. In future work, we aim to extend this procedure model to prescribe how a long-term management, i.e., changes to ontology or process model, can be handled.

\section{Conclusion}

So, OBPM: will we live to see it? As a main takeaway, while the primary focus in research has been on R6 (ontology-based reasoning, cf. Section \ref{sec:analysis}), the proposed requirements show that there are many more components still needed for implementing OBPM, especially in regard to initializing and evolving models. To this aim, our distilled research agenda provides a guideline for developing these needed results, and -- on a positive note -- shows that the field of OBPM offers many opportunities for future research. If we would have to propose one ``paradigm change", we would opt for less works such as ``yet another approach for using ontology-based process models" and more awareness for the initial creation and long-term management of ontology-based process models themselves. Also, the number of works that investigate the actual cognitive effects and usability of process models is alarmingly low, which urgently has to be addressed to ensure that academic results and company use cases are aligned. Should these requirements be met, OBPM can and will be successfully implemented.

We identify the following limitations. First, this review was limited to ontologies formalized with description logics. While this is an important fragment of OBPM, further works could also investigate the role of foundational ontologies in regard to OBPM. Second, the requirements identified in this work were derived from a design science perspective, i.e., they are requirements needed for an initial implementation. While we could already use these initial requirements to show limitations in the state of the art, these requirements were not further evaluated. In future works, we will therefore evaluate these requirements with domain experts. Last, this work investigates the state of the art of scientific contributions. In future work, we will further investigate industrial applications and solutions and consolidate them with the findings of this work.

\subsection*{Acknowledgements}
We thank Hagen Kerber for his preliminary research for this report. We also thank Silvano Colombo Tosatto and colleagues for letting us borrow their title idea \cite{tosatto2019checking}.

%\section*{References}
%Due to excessive length, the references can be found in the %supplementary material: TODO URL

\bibliography{references}

\end{document}